\documentclass[11pt]{article}
\usepackage[preprint]{acl}
\usepackage{times}
\usepackage{latexsym}
\usepackage[T1]{fontenc}
\usepackage[utf8]{inputenc}
\usepackage{microtype}
\usepackage{inconsolata}
\usepackage{amsmath,amssymb}
\usepackage{booktabs,multirow,colortbl}
\usepackage{graphicx}

\newcommand{\method}{\mbox{\textsc{Recap}}}
\newcommand{\U}{\mathcal{U}}
\newcommand{\C}{\mathcal{C}}
\newcommand{\G}{\mathcal{G}}
\newcommand{\speed}{\!\times}
\definecolor{recaprow}{HTML}{E8E6FF}

\title{RECAP: Context-Aware Prediction Recycling with
Request-Local and Persistent Memories for Multimodal
Speculative Decoding}

\author{
    Hengjie Zhu\textsuperscript{1,2},
    Dayan Wu\textsuperscript{1}\thanks{Corresponding author.},
    Xinze Liu\textsuperscript{1,2},\\
    Peng Fu\textsuperscript{1},
    Zheng Lin\textsuperscript{1},
    Weiping Wang\textsuperscript{1}\\[6pt]
    \small\textsuperscript{1}Institute of Information Engineering,
    Chinese Academy of Sciences\\
    \small\textsuperscript{2}School of Cyber Security,
    University of Chinese Academy of Sciences\\
    \small\texttt{\{zhuhengjie, wudayan, liuxinze, fupeng,
    linzheng, wangweiping\}@iie.ac.cn}
}

\begin{document}
\maketitle
\begin{abstract}
Recycling a target model's own predictions offers a training-free source of proposals for speculative decoding. In vision-language models, however, existing recycling indexes predictions by the last token alone, conflating predictions computed under different images and prefixes; we find that such reuse fails most often at image-sensitive decoding states. Longer keys reduce this aliasing but match fewer states. We introduce \method, which stores target predictions under one-, two-, and three-token keys: contextual memories live only for the current request and are thus implicitly conditioned on its image, while token-level memory persists across requests to retain coverage. At each lookup, entropy-confidence weighting fuses the retrieved predictions according to their key specificity and concentration. Best-first search composes the fused transitions into a candidate tree, supplemented by suffix-matched continuations, and one target pass verifies the tree under greedy decoding. Without training or an auxiliary draft model, \method\ achieves mean speedups over autoregressive decoding of $1.66\speed$ and $1.56\speed$ across seven single-turn benchmarks on Qwen2.5-VL-7B and LLaVA-1.6-7B, exceeding the strongest training-free baseline, Token Recycling ($1.51\speed$ and $1.37\speed$). Our implementation is available in https://github.com/Senmo996/RECAP-official-code.
\end{abstract}

\begin{figure}[!t]
\centering
\includegraphics[width=\columnwidth]{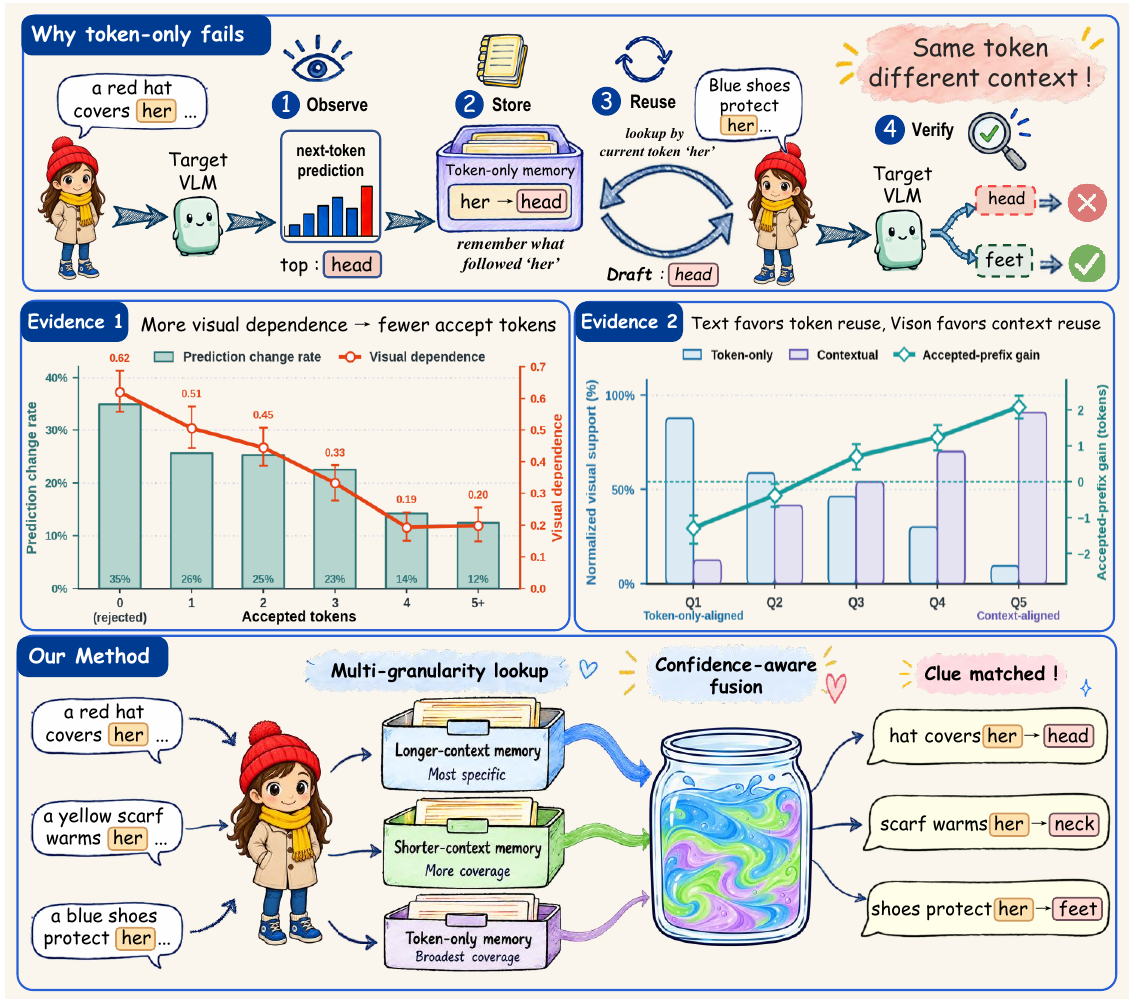}
\caption{The role of context in prediction recycling. Top: a cached \emph{her}$\to$\emph{head} transition fails when reused after a different prefix. Middle: shorter accepted paths accompany stronger root image sensitivity (left), and source preference varies with candidate visual alignment (right). Bottom: \method\ retrieves and fuses predictions at multiple contextual resolutions.}
\label{fig:intro}
\end{figure}

\section{Introduction}
\label{sec:intro}

Vision-language models (VLMs) generate text one token at a time, so every output token requires a sequential target-model pass. Speculative decoding reduces the number of sequential passes by proposing several tokens and verifying them in one parallel pass \citep{pmlr-v202-leviathan23a,chen2023acceleratinglargelanguagemodel}. Multimodal methods usually obtain proposals from a separately trained drafter \citep{hu2025dreamdraftingrefinedtarget,kang2025vispecacceleratingvisionlanguagemodels}, adding training and serving cost for every target model. Tree verification offers a cheaper source: each pass computes full next-token distributions at tens of candidate nodes yet commits only a few tokens, discarding the rest. Prediction recycling stores these target predictions and reuses them as later proposals \citep{luo-etal-2025-turning}, requiring neither a draft model nor training.

A recycled prediction is valid for the condition under which the target computed it: the image and the textual prefix. Existing recycling indexes predictions by the last token alone; Token Recycling, for example, keeps one Top-$k$ prediction row per vocabulary token in an adjacency matrix that every verification pass overwrites and that is retained across requests \citep{luo-etal-2025-turning}. Such a key merges all conditions that end in the same token, a mismatch we call \emph{condition aliasing}. Along the prefix, \emph{her}$\to$\emph{head}, cached after \emph{``a red hat covers her''}, is reused after \emph{``blue shoes protect her''}, where the target selects \emph{feet} (Figure~\ref{fig:intro}). Along the image, a prefix such as \emph{``the sign reads''} must continue with whatever text the current image shows. Reuse also fails at image-sensitive states: in counterfactual diagnostics on $700$ Qwen2.5-VL images, replacing the image changes the root prediction for $34.9\%$ of fully rejected drafts but for only $12.4\%$ of drafts that accept five or more tokens.

Longer keys reduce aliasing but recur less often: they recall the target token more reliably when they match, yet match far fewer decoding states. We call this tension the \emph{specificity--coverage trade-off}. Suffix-matching methods such as Prompt Lookup Decoding and SAM Decoding condition on longer contexts \citep{saxena2023prompt,hu-etal-2025-sam}, but each match copies a single observed continuation rather than combining target predictions obtained under several context lengths.

The two parts of the condition call for separate controls. \emph{Key length distinguishes prefixes.} Longer keys separate prefixes that a token key aliases, while shorter keys remain necessary for coverage. Neither resolution dominates: whether contextual or token-level candidates yield longer accepted prefixes depends on which carries more visual support (Appendix~\ref{app:intro-diagnostics}), so candidate selection should combine resolutions. \emph{Memory lifetime distinguishes images.} A lexical key cannot encode the image, but the image is fixed within a request, so request-local memory conditions its predictions on that image implicitly. Token transitions, in contrast, recur across requests, so token-level memory remains useful when retained across a stream.

\begin{figure*}[t]
\centering
\includegraphics[width=\textwidth]{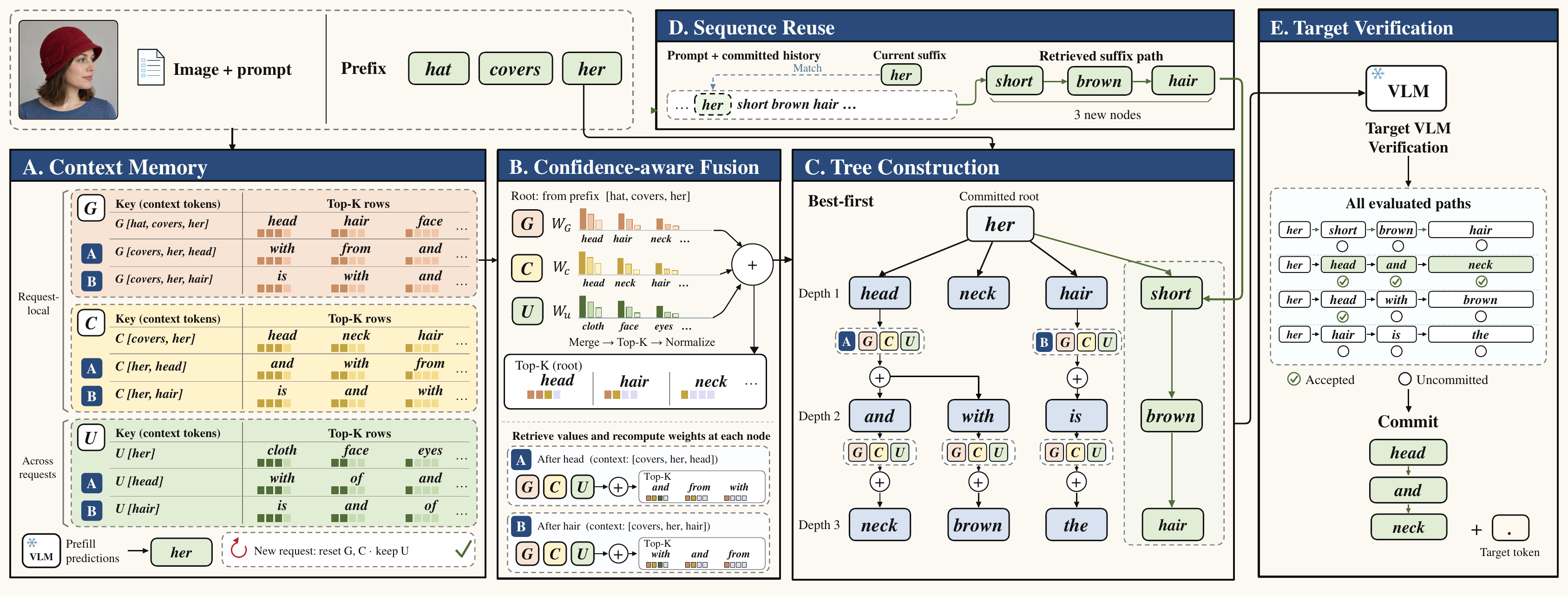}
\caption{The \method\ framework. (A) Three-, two-, and one-token keys retrieve Top-$K$ prediction rows from $\G$, $\C$, and $\U$; only $\U$ survives request boundaries. (B) Entropy-confidence weighting fuses the retrieved rows under specificity priors and normalizes the retained candidates. (C) Best-first tree search repeats retrieval and fusion under each node's path context. (D) Suffix retrieval inserts a contiguous continuation from the prompt and committed history. (E) One packed target pass commits the matching path plus one target-selected token. All evaluated nodes refresh prediction memory; only committed tokens extend sequence history. Tokens and scores are illustrative.}
\label{fig:overview}
\end{figure*}

We propose \method\ (\emph{REcycling Context-Aware Predictions}), a training-free framework built on these two controls (Figure~\ref{fig:overview}). It stores target predictions under one-, two-, and three-token keys; the two contextual maps live for a single request, whereas the token-level map persists across requests. Because a specific row may be diffuse while a shorter row is sharply peaked, \method\ does not simply prefer the longest key: it weights each retrieved row by a specificity prior scaled by an entropy-based confidence and mixes the rows, reinforcing shared candidates while retaining confident alternatives. Best-first search composes the fused transitions into a candidate tree, and a suffix automaton adds short continuations from the prompt and committed history. One target pass verifies the tree, commits the longest matching path, and writes the predictions of every evaluated node back to memory.

Our contributions are as follows:
\begin{itemize}
\item We identify \emph{condition aliasing} in multimodal prediction recycling and show that token-indexed reuse fails most often at image-sensitive states.
\item We propose \method, a training-free framework that distinguishes prefixes through multi-length keys with entropy-confidence fusion and images through request-local and persistent memories.
\item \method\ achieves mean speedups of $1.66\speed$ and $1.56\speed$ on Qwen2.5-VL-7B and LLaVA-1.6-7B across seven single-turn benchmarks, and $1.89\speed$ and $1.59\speed$ on MMSpec, exceeding the strongest training-free baseline in mean speedup.
\end{itemize}

\section{Related Work}
\label{sec:related}

\paragraph{Speculative decoding.}
Speculative decoding separates proposal generation from parallel target verification \citep{pmlr-v202-leviathan23a,chen2023acceleratinglargelanguagemodel}. Learned proposers exploit target representations: Medusa adds future-token heads with tree verification \citep{pmlr-v235-cai24b}, and Hydra models dependencies between head predictions \citep{ankner2024hydrasequentiallydependentdraftheads}. EAGLE drafts autoregressively in feature space \citep{pmlr-v235-li24bt}, EAGLE-2 adapts draft trees to context-dependent confidence \citep{li-etal-2024-eagle}, and EAGLE-3 combines multi-layer features with direct token prediction \citep{NEURIPS2025_c7b5a35e}. \method\ keeps the propose--verify interface but drafts from memory of predictions the target has already evaluated.

\paragraph{Multimodal speculative decoding.}
Visual evidence adds a conditioning requirement to multimodal drafting: a drafter must propose tokens that are consistent with both the textual context and the visual input. Existing approaches address this challenge through different mechanisms. MSD separates text and visual token processing and trains a drafter with a text-to-multimodal curriculum \citep{lin2025speculativedecodingreimaginedmultimodal}; DREAM transfers refined target features through cross-attention \citep{hu2025dreamdraftingrefinedtarget}; ViSpec adds a vision adaptor to drafting \citep{kang2025vispecacceleratingvisionlanguagemodels}; SpecVLM combines EAGLE-style drafting, adaptive visual compression, and online-logit distillation \citep{huang2025specvlmfastspeculativedecoding}. Together, these methods explore how visual information can be incorporated into a learned drafting process. MMSpec evaluates such approaches across multimodal task categories \citep{shen2026mmspecbenchmarkingspeculativedecoding}. \method\ instead obtains visual conditioning from cached target predictions and request-local memory, reusing information produced under the current multimodal context without training a multimodal proposer.

\paragraph{Training-free speculative decoding.}
Training-free methods derive proposals from existing sequences and target computations. Prompt Lookup Decoding reuses repeated prompt substrings \citep{saxena2023prompt}; REST retrieves suffix-matched continuations from an external datastore \citep{he-etal-2024-rest}; SAM Decoding finds longest-suffix matches in static corpora and generation history \citep{hu-etal-2025-sam}; SuffixDecoding builds frequency-scored trees from current and previous outputs \citep{NEURIPS2025_b7aea253}. Lookahead generates and verifies $n$-grams through parallel iteration \citep{pmlr-v235-fu24a}, and Token Recycling forms draft trees from token-indexed predictions \citep{luo-etal-2025-turning}. Related memory mechanisms include neural caches \citep{grave2016improvingneurallanguagemodels}, $k$NN-LMs \citep{khandelwal2020generalizationmemorizationnearestneighbor}, and hierarchical language models that support sparse longer contexts with shorter-context distributions \citep{teh-2006-hierarchical}. \method\ combines multi-resolution lexical keys with request-local and persistent scopes, while the target retains control of commitment.

\section{Context-Aware Prediction Recycling}
\label{sec:method}

\subsection{Motivation}
\label{sec:motivation}
We examine how reuse depends on visual context through the $700$-image counterfactual diagnostics in Figure~\ref{fig:intro} (Appendix~\ref{app:intro-diagnostics}). With each text prefix fixed, target predictions are recomputed under mean-image and matched wrong-image controls, and results are averaged equally across benchmarks.

\emph{Evidence~1} relates draft acceptance to root sensitivity. With iterations grouped by accepted length, bars show root Top-$1$ changes under either control, and the curve gives the smaller of the two decreases in the original root token's log probability. Rejected drafts score higher on both measures than $5+$-token acceptances.

\emph{Evidence~2} compares sources at equal candidate budgets. With the prefix and candidate set fixed, visual support $E_s$ is the smaller of the two reductions, under the image controls, in the log of the total target probability assigned to source $s$'s candidates. Relative support $E_{GC}-E_U$ between contextual ($\G/\C$) and token-only ($\U$) candidates defines within-benchmark quintiles from token-only-aligned Q1 to context-aligned Q5. Bars show softmax-normalized support shares, and the curve shows contextual minus token-only accepted-prefix length. Its sign reversal indicates that the better source changes with candidate alignment, motivating memory that retains multiple context resolutions and fuses their evidence.

\subsection{Problem Formulation}
\label{sec:formulation}

Let $I$ be the image, $p$ the prompt, and $y_{<t}$ the generated prefix, with textual history $h=(p,y_{<t})$. Greedy decoding with the VLM $P_\theta$ over vocabulary $\mathcal V$ selects
\begin{equation}
 y_t=\arg\max_{z\in\mathcal V}P_\theta(z\mid I,h),
 \label{eq:target-decoding}
\end{equation}

Speculative decoding verifies a proposed sequence or tree in parallel, commits the longest prefix matching Equation~\ref{eq:target-decoding}, and emits one additional target-selected token \citep{pmlr-v202-leviathan23a}, so accepting $a$ draft tokens yields $a+1$ output tokens per pass. \method\ preserves this commitment rule while building proposals from recycled target predictions.

\subsection{Context-Stratified Prediction Memory}
\label{sec:memory}

At each evaluated position, the target assigns logit $\ell_z$ to token $z$. We retain the $K$ highest-scoring tokens as $\mathcal S_K$. A stored \emph{prediction row} consists of $\mathcal S_K$ and its normalized distribution
\begin{equation}
 \widehat P(z)=
 \frac{\exp(\ell_z)}{\sum_{v\in\mathcal S_K}\exp(\ell_v)},
 \quad z\in\mathcal S_K,
 \label{eq:row}
\end{equation}
with zero mass on all other tokens. Each row thus inherits the conditioning of its source image and prefix.

Rows are addressed by the last one, two, or three tokens of $h$. For suffix $(u,v,x)$, these keys retrieve $\U[x]$, $\C[v,x]$, and $\G[u,v,x]$. Thus, \emph{covers her}, \emph{warms her}, and \emph{protect her} share the token key \emph{her} but access distinct contextual rows (Figure~\ref{fig:intro}), reducing condition aliasing.

\paragraph{Context refinement.}
Let $H_j$ be the final $j$ history tokens, and let $R_K^*(H_j)$ denote the maximum expected recall of the target's greedy token for selectors using only $H_j$ and at most $K$ candidates. Since $H_j$ is a deterministic function of $H_{j+1}$,
\begin{equation}
 R_K^*(H_j)\le R_K^*(H_{j+1}),\quad j=1,2.
 \label{eq:context-refinement}
\end{equation}
A finer-context selector can reproduce every coarse-context decision, but in a finite memory finer keys are observed less often; retaining shorter rows supplies coverage where longer keys are missing, as in hierarchical language models \citep{teh-2006-hierarchical}. Appendix~\ref{app:theory} gives the proof.

\paragraph{Memory scope.}
Contextual maps $\C$ and $\G$ reset per request, making the current image and prompt implicit parts of the retrieval condition, whereas $\U$ persists within an inference stream and supplies earlier transitions when local memory is sparse. Prefill seeds $\U$, tree verification refreshes all three maps, and each key retains its most recently assigned row.

\begin{table*}[t]
\centering
\footnotesize
\renewcommand{\arraystretch}{1.10}
\setlength{\tabcolsep}{2.3pt}
\begin{tabular}{@{}cc*{16}{>{\raggedleft\arraybackslash}p{1.85em}}@{}}
\toprule
\multirow{2}{*}{Models} & \multirow{2}{*}{Methods} & \multicolumn{2}{c}{MMT} & \multicolumn{2}{c}{SEED} & \multicolumn{2}{c}{SciQA} & \multicolumn{2}{c}{OCR} & \multicolumn{2}{c}{ChartQA} & \multicolumn{2}{c}{MathVista} & \multicolumn{2}{c}{TextVQA} & \multicolumn{2}{c}{Average} \\
\cmidrule(lr){3-4}\cmidrule(lr){5-6}\cmidrule(lr){7-8}\cmidrule(lr){9-10}\cmidrule(lr){11-12}\cmidrule(lr){13-14}\cmidrule(lr){15-16}\cmidrule(lr){17-18}
 & & $S$ & $\tau$ & $S$ & $\tau$ & $S$ & $\tau$ & $S$ & $\tau$ & $S$ & $\tau$ & $S$ & $\tau$ & $S$ & $\tau$ & $S$ & $\tau$ \\
\midrule
\multirow{7}{*}{\shortstack{Qwen2.5-VL\\7B-Instruct}} & PLD & 1.16 & 1.19 & 0.89 & 1.05 & 1.24 & 1.13 & 1.22 & 1.20 & 1.14 & 1.25 & 1.49 & 1.22 & 0.97 & 1.09 & 1.16 & 1.16 \\
 & Lookahead & 1.15 & 1.60 & 0.93 & 1.35 & 1.44 & 2.14 & 1.11 & 1.62 & 1.10 & 1.62 & 1.20 & 1.60 & 0.94 & 1.26 & 1.12 & 1.60 \\
 & Tree Recycling & 1.20 & 1.47 & 1.07 & 1.49 & 1.23 & 1.54 & 1.29 & 1.68 & 1.27 & 1.72 & 1.41 & 1.70 & 1.06 & 1.41 & 1.22 & 1.57 \\
 & Token Recycling & 1.59 & 2.57 & 1.25 & 2.58 & 1.60 & 2.60 & 1.51 & 2.82 & 1.54 & 2.89 & 1.86 & 2.84 & 1.21 & 2.63 & 1.51 & 2.71 \\
 & SAM & 1.11 & 1.24 & 1.06 & 1.34 & 1.15 & 1.36 & 1.20 & 1.33 & 1.23 & 1.57 & 1.21 & 1.38 & 1.05 & 1.26 & 1.14 & 1.35 \\
 & SAM+ViSkip & 1.04 & 1.25 & 0.99 & 1.27 & 1.07 & 1.29 & 1.04 & 1.28 & 1.16 & 1.50 & 1.14 & 1.36 & 0.99 & 1.22 & 1.06 & 1.31 \\
\rowcolor{recaprow}
\cellcolor{white} & \textbf{\method} & \textbf{1.69} & \textbf{2.81} & \textbf{1.27} & \textbf{2.78} & \textbf{1.72} & \textbf{2.84} & \textbf{1.67} & \textbf{3.31} & \textbf{1.75} & \textbf{3.30} & \textbf{2.19} & \textbf{3.27} & \textbf{1.34} & \textbf{2.81} & \textbf{1.66} & \textbf{3.02} \\
\midrule
\multirow{7}{*}{\shortstack{LLaVA-1.6\\Vicuna-7B}} & PLD & 1.01 & 1.34 & 0.91 & 1.14 & 1.09 & 1.23 & 1.28 & 1.53 & 1.02 & 1.34 & 1.19 & 1.40 & 0.89 & 1.11 & 1.06 & 1.30 \\
 & Lookahead & 0.80 & 1.74 & 0.70 & 1.48 & 0.94 & 2.11 & 0.91 & 2.11 & 0.74 & 1.66 & 0.79 & 1.70 & 0.64 & 1.31 & 0.79 & 1.73 \\
 & Tree Recycling & 1.23 & 1.48 & 1.15 & 1.49 & 1.22 & 1.43 & 1.40 & 1.82 & 1.28 & 1.60 & 1.37 & 1.69 & 1.13 & 1.34 & 1.25 & 1.55 \\
 & Token Recycling & 1.40 & 2.48 & 1.24 & 2.49 & 1.36 & 2.47 & 1.49 & 2.69 & 1.42 & 2.66 & 1.49 & 2.65 & 1.23 & 2.41 & 1.37 & 2.55 \\
 & SAM & 1.13 & 1.26 & 1.11 & 1.34 & 1.12 & 1.24 & 1.24 & 1.45 & 1.18 & 1.36 & 1.23 & 1.40 & 1.07 & 1.19 & 1.16 & 1.32 \\
 & SAM+ViSkip & 1.10 & 1.26 & 1.07 & 1.34 & 1.09 & 1.24 & 1.20 & 1.45 & 1.15 & 1.36 & 1.20 & 1.40 & 1.04 & 1.19 & 1.12 & 1.32 \\
\rowcolor{recaprow}
\cellcolor{white} & \textbf{\method} & \textbf{1.63} & \textbf{2.77} & \textbf{1.31} & \textbf{2.82} & \textbf{1.56} & \textbf{2.70} & \textbf{1.81} & \textbf{3.40} & \textbf{1.58} & \textbf{3.06} & \textbf{1.71} & \textbf{3.11} & \textbf{1.29} & \textbf{2.59} & \textbf{1.56} & \textbf{2.92} \\
\bottomrule
\end{tabular}
\caption{Speedup $S$ ($\times$) and acceptance length $\tau$ across seven single-turn benchmarks at temperature $0$. SciQA denotes ScienceQA and OCR denotes OCRBench. $S$ averages paired generation-throughput ratios against autoregressive decoding; $\tau$ averages accepted draft tokens plus one target-selected token. Higher is better for both metrics.}
\label{tab:main}
\end{table*}

\begin{table*}[t]
\centering
\footnotesize
\renewcommand{\arraystretch}{1.10}
\setlength{\tabcolsep}{3.5pt}
\begin{tabular}{cc*{7}{cc}}
\toprule
\multirow{2}{*}{Models} & \multirow{2}{*}{Methods} & \multicolumn{2}{c}{General VQA} & \multicolumn{2}{c}{Text VQA} & \multicolumn{2}{c}{Captioning} & \multicolumn{2}{c}{Charts} & \multicolumn{2}{c}{Reasoning} & \multicolumn{2}{c}{Multi-turn} & \multicolumn{2}{c}{Overall} \\
\cmidrule(lr){3-4}\cmidrule(lr){5-6}\cmidrule(lr){7-8}\cmidrule(lr){9-10}\cmidrule(lr){11-12}\cmidrule(lr){13-14}\cmidrule(lr){15-16}
 & & $S$ & $\tau$ & $S$ & $\tau$ & $S$ & $\tau$ & $S$ & $\tau$ & $S$ & $\tau$ & $S$ & $\tau$ & $S$ & $\tau$ \\
\midrule
\multirow{7}{*}{\shortstack{Qwen2.5-VL\\7B-Instruct}} & PLD & 1.06 & 1.05 & 1.21 & 1.13 & 1.14 & 1.15 & 1.20 & 1.31 & 1.28 & 1.34 & 1.34 & 1.27 & 1.23 & 1.22 \\
 & Lookahead & 1.11 & 1.37 & 1.04 & 1.29 & 1.44 & 1.64 & 1.21 & 1.77 & 1.91 & 2.28 & 1.39 & 1.80 & 1.36 & 1.71 \\
 & Tree Recycling & 1.12 & 1.40 & 1.10 & 1.32 & 1.22 & 1.37 & 1.21 & 1.64 & 1.38 & 1.66 & 1.25 & 1.56 & 1.22 & 1.51 \\
 & Token Recycling & 1.51 & 2.57 & \textbf{1.31} & \textbf{2.49} & 1.80 & 2.51 & 1.38 & 2.60 & 1.77 & 2.64 & 1.57 & 2.60 & 1.56 & 2.57 \\
 & SAM & 1.05 & 1.14 & 1.13 & 1.13 & 1.03 & 1.07 & 1.14 & 1.54 & 1.18 & 1.31 & 1.08 & 1.23 & 1.10 & 1.24 \\
 & SAM+ViSkip & 0.97 & 1.16 & 1.00 & 1.12 & 0.95 & 1.06 & 1.06 & 1.55 & 1.07 & 1.29 & 0.99 & 1.24 & 1.00 & 1.24 \\
\rowcolor{recaprow}
\cellcolor{white} & \textbf{\method} & \textbf{1.77} & \textbf{2.69} & 1.26 & \textbf{2.49} & \textbf{2.00} & \textbf{2.60} & \textbf{1.86} & \textbf{3.02} & \textbf{2.16} & \textbf{2.97} & \textbf{2.09} & \textbf{2.80} & \textbf{1.89} & \textbf{2.77} \\
\midrule
\multirow{7}{*}{\shortstack{LLaVA-1.6\\Vicuna-7B}} & PLD & 1.03 & 1.10 & 1.06 & 1.08 & 1.09 & 1.17 & 1.21 & 1.54 & 1.37 & 1.79 & 1.33 & 1.82 & 1.21 & 1.48 \\
 & Lookahead & 0.93 & 1.41 & 0.85 & 1.25 & 0.87 & 1.34 & 1.02 & 1.66 & 1.03 & 1.72 & 0.90 & 1.43 & 0.92 & 1.46 \\
 & Tree Recycling & 1.12 & 1.43 & 1.17 & 1.40 & 1.19 & 1.34 & 1.34 & 1.70 & 1.37 & 1.77 & 1.35 & 1.64 & 1.27 & 1.56 \\
 & Token Recycling & 1.21 & 2.42 & 1.35 & 2.48 & 1.50 & 2.50 & 1.43 & 2.58 & 1.46 & 2.57 & 1.53 & 2.65 & 1.43 & 2.55 \\
 & SAM & 1.07 & 1.23 & 1.06 & 1.16 & 1.04 & 1.09 & 1.24 & 1.46 & 1.22 & 1.49 & 1.16 & 1.28 & 1.14 & 1.29 \\
 & SAM+ViSkip & 1.04 & 1.23 & 1.04 & 1.16 & 1.02 & 1.09 & 1.21 & 1.46 & 1.20 & 1.49 & 1.14 & 1.29 & 1.12 & 1.29 \\
\rowcolor{recaprow}
\cellcolor{white} & \textbf{\method} & \textbf{1.26} & \textbf{2.60} & \textbf{1.44} & \textbf{2.64} & \textbf{1.65} & \textbf{2.63} & \textbf{1.65} & \textbf{3.10} & \textbf{1.67} & \textbf{3.09} & \textbf{1.73} & \textbf{2.93} & \textbf{1.59} & \textbf{2.85} \\
\bottomrule
\end{tabular}
\caption{MMSpec speedup $S$ ($\times$) and acceptance length $\tau$ at temperature $0$, with the metric definitions of Table~\ref{tab:main}. Each category contains $100$ inputs; multi-turn conversations contribute $223$ generation turns, giving $723$ total turns. Overall values are turn-weighted averages of both metrics, rather than unweighted category means.}
\label{tab:mmspec}
\end{table*}

\subsection{Entropy-Confidence-Weighted Fusion}
\label{sec:fusion}

Let $(M_1,\ldots,M_r)$ denote the rows available for history $h$, ordered from most to least specific ($\G$, $\C$, $\U$). \method\ fuses them by \emph{entropy-confidence weighting}: each row's weight is the product of a specificity prior, with $\alpha_1\ge\alpha_2\ge\alpha_3>0$, and an entropy-based confidence, defined as the row's inverse effective support:
\begin{equation}
 \begin{aligned}
 c_j(h)&=\exp\!\left[-H\!\left(\widehat P_{M_j}\right)\right],\\
 w_j(h)&=\frac{\alpha_j c_j(h)}{\sum_{i=1}^{r}\alpha_i c_i(h)},
 \end{aligned}
 \label{eq:confidence-weights}
\end{equation}
where $H(P)=-\sum_z P(z)\log P(z)$. The prior favors specific matches, and confidence favors concentrated predictions. Priors follow specificity rank among available rows: if only $\C$ and $\U$ match, they receive $\alpha_1$ and $\alpha_2$. Weights are recomputed at every tree node without additional target evaluations or learned parameters. Confidence measures concentration under the source condition, not direct alignment with the current image.

The normalized weights define the mixture
\begin{equation}
 m(z\mid h)=\sum_{j=1}^{r}w_j(h)\widehat P_{M_j}(z).
 \label{eq:fusion}
\end{equation}
We bound branching by selecting the Top-$K$ tokens $\mathcal S(h)$, using the same $K$ as row storage, and renormalizing:
\begin{equation}
 q(z\mid h)=\frac{m(z\mid h)}{\sum_{v\in\mathcal S(h)}m(v\mid h)},
 \quad z\in\mathcal S(h),
 \label{eq:proposal}
\end{equation}
with all other probabilities set to zero. A single retrieved row recovers ordinary lookup, and no match terminates expansion. Setting $c_j(h)=1$ for all available rows yields the fixed-prior special case, the \emph{w/o confidence} configuration analyzed in Appendix~\ref{app:adaptive-fusion}.

\paragraph{Weighted KL center.}
For fixed $h$, the weights do not depend on the candidate distribution $\rho$, so the mixture is the weighted Kullback--Leibler (KL) center \citep{Abbas_2009}:
\begin{equation}
 m(\cdot\mid h)=\operatorname*{arg\,min}_{\rho\in\Delta(\mathcal V)}
 \sum_{j=1}^{r}w_j(h)D_{\mathrm{KL}}\bigl(\widehat P_{M_j}\,\|\,\rho\bigr),
 \label{eq:fusion-centroid}
\end{equation}
where $\Delta(\mathcal V)$ is the probability simplex. Top-$K$ retention maximizes the preserved mixture mass, and the normalized proposal minimizes $D_{\mathrm{KL}}(q\|m)$ over distributions supported on at most $K$ tokens (Appendix~\ref{app:theory}).

\subsection{Tree Construction and Verification}
\label{sec:suffix}
\label{sec:verification}

At every node, \method\ retrieves and fuses rows using the established prefix together with proposed ancestors. Best-first construction orders paths $\pi=(z_1,\ldots,z_d)$ by the score
\begin{equation}
 \phi(\pi)=\prod_{j=1}^{d}\eta_j q(z_j\mid h_j),
 \label{eq:tree}
\end{equation}
where $h_j=(h,z_{<j})$ supplies the path context and $\eta_j$ controls depth-dependent expansion, until the node or depth budget is reached. Composing local predictions yields candidate sequences never stored as complete continuations.

A suffix automaton over the prompt and committed history retrieves the continuation following a match to the current suffix \citep{hu-etal-2025-sam} and inserts it into the tree by sharing common prefixes, preserving observed sequence dependencies without modifying the fused transition scores.

The target verifies the joint tree in one packed pass, with ancestral attention confining each node to the established prefix and its own path. The longest path matching Equation~\ref{eq:target-decoding} is committed together with one target-selected token; if no candidate agrees, the target directly supplies the next token.

Predictions from \emph{all evaluated nodes}, including uncommitted branches, update $\U$, $\C$, and $\G$ under their path contexts, whereas only committed tokens enter the suffix automaton. An uncommitted branch thus contributes a target prediction valid under its hypothetical prefix, which may become useful when a compatible context recurs. Appendix~\ref{app:implementation} provides implementation settings and hyperparameters.

\section{Experimental Evaluation}
\label{sec:experiments}

\subsection{Setup}
\label{sec:setup}

\paragraph{Models and workloads.}
Evaluation uses Qwen2.5-VL-7B-Instruct \citep{bai2025qwen25vltechnicalreport} and LLaVA-1.6 Vicuna-7B \citep{liu2024llavanext} on an NVIDIA A100-SXM4-80GB GPU. Generation is greedy at batch size one, with at most $200$ new tokens. We fix $1{,}000$ inputs for each single-turn benchmark: MMT-Bench \citep{ying2024mmtbenchcomprehensivemultimodalbenchmark}, SEED-Bench \citep{li2023seedbenchbenchmarkingmultimodalllms}, ScienceQA \citep{lu2022learnexplainmultimodalreasoning}, OCRBench \citep{liu2024ocrbenchhiddenmysteryocr}, ChartQA \citep{masry-etal-2022-chartqa}, MathVista \citep{lu2024mathvistaevaluatingmathematicalreasoning}, and TextVQA \citep{Singh_2019_CVPR}. MMSpec adds $600$ inputs from six categories, extending coverage to multi-turn conversations \citep{shen2026mmspecbenchmarkingspeculativedecoding}. For the main results, each benchmark, including MMSpec, forms one inference stream: persistent $\U$ is cleared after warmup and before each benchmark, then carried across that benchmark's inputs, whereas $\C$ and $\G$ reset per request.

\paragraph{Baselines.}
We compare six draft-model-free approaches: Prompt Lookup Decoding (PLD; \citealp{saxena2023prompt}), Lookahead Decoding \citep{pmlr-v235-fu24a}, Token Recycling \citep{luo-etal-2025-turning}, Tree Recycling, SAM Decoding \citep{hu-etal-2025-sam}, and SAM+ViSkip \citep{shen2026mmspecbenchmarkingspeculativedecoding}. Token Recycling and Tree Recycling run token-indexed recycling \citep{luo-etal-2025-turning} on the same prefill and verification path as \method. Token Recycling follows the published configuration: a Top-$8$ adjacency matrix retained across requests drafts a static $80$-node tree, and the matrix is cleared after warmup and at each benchmark boundary, as is $\U$. Tree Recycling is a lighter variant that expands a smaller static tree from a request-local Top-$4$ matrix. All methods use fixed input manifests and an autoregressive reference from the same backbone. Their settings are listed in Appendix~\ref{app:implementation}.

\paragraph{Metrics.}
We report \emph{speedup} $S=\frac{1}{N}\sum_{i=1}^{N}r_i^m$ as the mean of paired throughput ratios, defined by
\begin{equation}
 r_i^m=\frac{n_i^m/T_i^m}{n_i^{\mathrm{AR}}/T_i^{\mathrm{AR}}}.
 \label{eq:speed}
\end{equation}
For each generation unit, $n_i^m$ and $T_i^m$ are the output-token count and total generation time for method $m$. The autoregressive reference is denoted $\mathrm{AR}$, and $N$ is the number of generation units.

Our \emph{acceptance length} $\tau$ includes the additional target-selected token:
\begin{equation}
 \tau=1+\frac{1}{N}\sum_{i=1}^{N}\bar a_i,
 \label{eq:acceptance}
\end{equation}
where $\bar a_i$ averages draft tokens accepted per verification iteration for unit $i$.

When target evaluations are exact, \method\ preserves greedy decoding without loss.

\subsection{Main Results}
\label{sec:mainresults}

\method\ leads every single-turn model--benchmark comparison in speedup and acceptance (Table~\ref{tab:main}). Qwen's mean speedup across seven benchmarks is $1.66\speed$, compared with $1.51\speed$ for Token Recycling, the strongest baseline by mean speedup. LLaVA's corresponding values are $1.56\speed$ and $1.37\speed$. The gains cover reasoning, chart understanding, and text-centric inputs: Qwen reaches $2.19\speed$ on MathVista and $1.75\speed$ on ChartQA, while LLaVA reaches $1.81\speed$ on OCRBench and $1.71\speed$ on MathVista. Alongside its consistent lead in $S$, \method\ has the highest $\tau$ throughout. Its mean acceptance lengths of $3.02$ and $2.92$ exceed Token Recycling's $2.71$ and $2.55$, the strongest baseline acceptance averages.

On MMSpec (Table~\ref{tab:mmspec}), \method\ reaches overall speedups of $1.89\speed$ on Qwen and $1.59\speed$ on LLaVA, exceeding Token Recycling, the strongest baseline on both models, at $1.56\speed$ and $1.43\speed$. \method\ leads in both speedup and acceptance length across categories except Qwen text VQA, where Token Recycling reaches $1.31\speed$ against $1.26\speed$, while both methods round to $\tau=2.49$. Overall acceptance lengths for \method\ are $2.77$ and $2.85$. Multi-turn conversation yields $2.09\speed$ on Qwen, second only to complex reasoning at $2.16\speed$, and $1.73\speed$ on LLaVA, its largest category speedup.

\begin{table}[t]
\centering
\small
\renewcommand{\arraystretch}{1.10}
\setlength{\tabcolsep}{4.5pt}
\begin{tabular}{llrrrrr}
\toprule
Model & Method & Params & GiB & $\tau$ & $S$ & AR \\
\midrule
 & DREAM & 607M & 1.13 & 3.68 & 1.26 & 41.9 \\
 & ViSpec & 451M & 0.84 & 3.74 & 1.05 & 61.2 \\
\rowcolor{recaprow}
\cellcolor{white}\multirow{-3}{*}{LLaVA} & \textbf{\method} & \textbf{0} & \textbf{0} & 2.92 & 1.56 & 29.9 \\
\midrule
 & ViSpec & 890M & 1.66 & 3.76 & 1.18 & 57.0 \\
\rowcolor{recaprow}
\cellcolor{white}\multirow{-2}{*}{Qwen} & \textbf{\method} & \textbf{0} & \textbf{0} & 3.02 & 1.66 & 39.4 \\
\bottomrule
\end{tabular}
\caption{Deployment cost and decoding performance on the seven single-turn benchmarks in Table~\ref{tab:main}. Params and GiB denote drafter parameter counts and 16-bit GPU weight storage; AR is mean autoregressive throughput (tokens/s). DREAM and ViSpec use released checkpoints with top-$4$, depth-$6$, $32$-node trees and a faster AR harness, so their $S$ is not directly comparable with that of \method.}
\label{tab:trained}
\end{table}

\paragraph{Comparison with trained drafters.}
Table~\ref{tab:trained} contrasts \method\ with released DREAM \citep{hu2025dreamdraftingrefinedtarget} and ViSpec \citep{kang2025vispecacceleratingvisionlanguagemodels} drafters on the same single-turn inputs. Trained drafters accept longer continuations, with mean $\tau$ of $3.68$--$3.76$ against $2.92$ and $3.02$ for \method, which reaches $78$--$80\%$ of their acceptance length. Their speedups of $1.05$--$1.26\speed$ fall below those of \method, but they were measured against autoregressive baselines of $41.9$--$61.2$ tokens/s rather than $29.9$--$39.4$ tokens/s, so this gap does not establish that \method\ decodes faster. Their acceptance advantage requires a training stage for each target model and adds $451$M--$890$M parameters, or $0.84$--$1.66$ GiB of 16-bit weights, to the serving GPU. \method\ adds no parameters or GPU-resident state; at the default budget, its proposal state occupies at most $8.70$ MiB of host memory (Table~\ref{tab:overhead-budgets}). It therefore transfers to a new VLM without data collection, training, or a drafter checkpoint.

\begin{figure*}[!t]
\centering
\includegraphics[width=\textwidth]{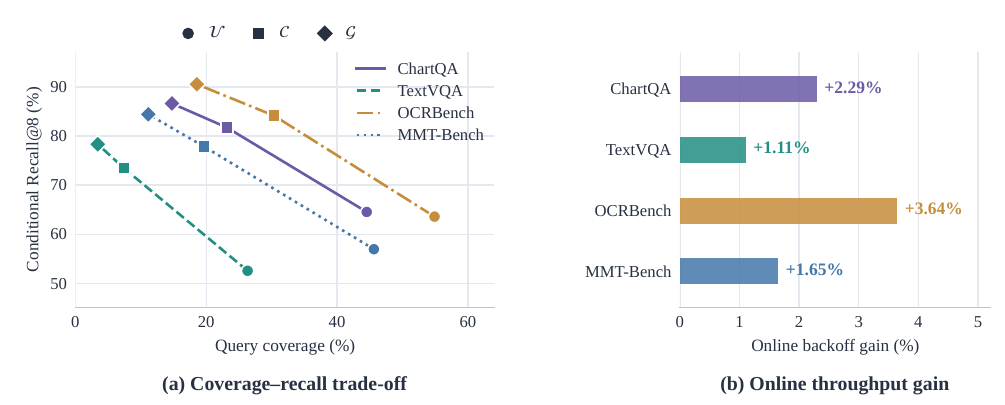}
\caption{The specificity--coverage trade-off and contextual backoff. (a) Query coverage and conditional Recall@$8$ across $23{,}811$ states from $400$ inputs. Each benchmark connects $\U$ (circle), $\C$ (square), and $\G$ (diamond). Finer keys raise conditional recall while reducing coverage. (b) Paired online throughput gains of $\U+\C$ backoff against $\U$.}
\label{fig:granularity}
\end{figure*}

\section{Analysis of Prediction Reuse}
\label{sec:analysis}

\subsection{Component Contributions}
\label{sec:ablation}

Matched-budget ablations test which parts of the design contribute to throughput (Table~\ref{tab:ablation}). The MMSpec subset comprises $120$ inputs, $20$ per category, expanding to $160$ generation turns, and is evaluated with Qwen2.5-VL-7B. Policies share a tree builder and a $63$-node transition budget. Reported gains average input-paired throughput changes after pooling each input's turn-level token counts and times. Except for the confidence comparison, these controls use the fixed-prior special case ($c_j=1$), so each comparison isolates one component.

\begin{table}[t]
\centering
\small
\renewcommand{\arraystretch}{1.10}
\setlength{\tabcolsep}{3pt}
\begin{tabular*}{\columnwidth}{@{\extracolsep{\fill}}lr@{}}
\toprule
Comparison & Gain (\%, $\uparrow$) \\
\midrule
\multicolumn{2}{l}{\emph{Candidate search}} \\
Best-first vs.\ fixed expansion & $+4.90$ \\
\midrule
\multicolumn{2}{l}{\emph{Memory design}} \\
Context backoff vs.\ unigram & $+3.03$ \\
Uniform fusion vs.\ backoff & $+0.75$ \\
Specificity priors vs.\ uniform fusion & $+2.80$ \\
Entropy confidence vs.\ fixed priors$^\dagger$ & $+0.73/{+}1.09$ \\
Persistent vs.\ reset $\U$ & $+5.42$ \\
\midrule
\multicolumn{2}{l}{\emph{Sequence reuse}} \\
Suffix insertion vs.\ none$^\dagger$ & $+0.61$ \\
\bottomrule
\end{tabular*}
\caption{Paired throughput effects of individual components. Unmarked rows use the $120$-input MMSpec subset with Qwen2.5-VL-7B; the persistence row averages five runs. $^\dagger$ marks four-dataset controls ($200$ inputs per model for confidence, reported as Qwen/LLaVA; $400$ inputs for suffix insertion). Backoff selects the most specific row available at a query.}
\label{tab:ablation}
\end{table}

Contextual backoff improves throughput over unigram recycling by $3.03\%$, and an independent repeat of the same $120$-input comparison yields $2.42\%$. Uniform fusion improves backoff by $0.75\%$, and specificity priors improve uniform fusion by $2.80\%$. Across five runs, retaining token memory across requests improves throughput over per-request reset by $5.42\%$ on average, ranging from $4.36\%$ to $7.18\%$. Request-stream controls are detailed in Appendix~\ref{app:scope}.

In four-dataset controls, entropy-confidence weighting adds a smaller gain over fixed priors: $0.73\%$ mean throughput on Qwen and $1.09\%$ on LLaVA. Confidence-based single-source gating decreases throughput by $0.66\%$ and $1.90\%$, respectively. The contrast supports combining source evidence in these settings while qualifying the size of the confidence benefit. Appendix~\ref{app:adaptive-fusion} details the adaptive rules and fixed-prior sensitivity.

Suffix complementarity is tested separately on $100$ inputs each from ChartQA, TextVQA, OCRBench, and MMT-Bench. Adding suffix insertion to fusion improves mean paired throughput by $0.61\%$, with dataset-specific changes from $-0.67\%$ to $+2.23\%$ (Table~\ref{tab:suffix}).

\subsection{Specificity Improves Matching; Backoff Retains Coverage}
\label{sec:granularity}

Replay comprises $23{,}811$ states from $400$ inputs, with $100$ each from ChartQA, TextVQA, OCRBench, and MMT-Bench. Retrieval begins with empty memories and accesses only rows from earlier states. \emph{Query coverage} is the fraction of states with a matching row. \emph{Conditional Recall@$8$} is recall of the reference next token among those covered states.

Stored rows preserve useful source-image information. To test this, we fix the query image, question, and prefix and vary the image used to compute earlier predictions. On jointly covered queries with eight candidates per source, matching-image rows outperform swapped-image rows in fused Recall@$8$ by $0.89$--$6.35$ percentage points across datasets (Appendix~\ref{app:additional}).

All four datasets show higher conditional recall and lower coverage with finer keys (Figure~\ref{fig:granularity}(a)). On OCRBench, $\U$ coverage is $54.9\%$ and $\G$ coverage is $18.6\%$, while recall rises from $63.6\%$ to $90.6\%$. In a prompt-seeded variant of this replay, restricting evaluation to states where both $\G$ and $\U$ match, $\G$ improves Recall@$8$ over $\U$ by $5.00$--$29.54$ percentage points across datasets.

At equal actual candidate counts, $\U+\C$ backoff also improves matched continuation length on every dataset, from $0.516$ to $0.709$ tokens on ChartQA and $0.632$ to $1.135$ on OCRBench (Appendix~\ref{app:additional}). This offline metric measures the longest tree prefix agreeing with the saved target continuation, excluding the root and target correction token.

Online controls yield $1.11$--$3.64\%$ paired throughput gains from $\U+\C$ backoff (Figure~\ref{fig:granularity}(b)). Adding $\G$ improves offline matching further on ChartQA and MMT-Bench, with only small throughput changes online. Context refinement therefore supplies more precise candidates, and backoff preserves the coverage needed for their practical reuse.

\subsection{Complementary Sequence Reuse}
\label{sec:sequence-analysis}

Suffix retrieval contributes sequence continuations beyond those formed by local transition composition. Starting with prompt-seeded $\U$, we compare fusion alone with suffix-augmented fusion at equal actual candidate counts. The added suffix increases mean matched length by $0.059$--$0.190$ tokens across four replay datasets (Appendix~\ref{app:additional}). The matched-budget result attributes the improvement to complementary sequence evidence.

Online throughput changes from suffix insertion range between $-0.67\%$ and $+2.23\%$ across the four datasets. Timing includes retrieval, insertion, and overlap handling. At the default budget, retrieval, fusion, tree construction, and packing together require $0.907$ ms per Qwen iteration and $1.121$ ms per LLaVA iteration. Including memory refresh, suffix-index updates, token commitment, and other host-side decoding work, all stages outside target verification take $1.778$ and $1.974$ ms, or $8.20\%$ and $5.61\%$ of decoding wall time (Appendix~\ref{app:overhead}).

\section{Conclusion}

\method\ addresses contextual specificity and reuse scope jointly in multimodal prediction recycling, fusing request-local contextual memories with persistent token-level memory through entropy-confidence weighting without training or an auxiliary draft model. It reaches mean generation-throughput speedups of $1.66\speed$ and $1.56\speed$ across seven single-turn benchmarks on Qwen2.5-VL-7B and LLaVA-1.6-7B, and $1.89\speed$ and $1.59\speed$ on MMSpec. Controlled analyses connect these gains to more precise contextual matching and reuse across request boundaries.

\section*{Limitations}

Our evidence covers greedy, batch-one generation with image-based inputs on two 7B VLMs. Larger models, concurrent serving, and video remain untested operating regimes. Default specificity priors, confidence weighting, and tree settings may yield different latency--memory trade-offs across hardware and request lengths. The reported efficiency concerns generation throughput and does not establish downstream task accuracy or application-level latency distributions.

\section*{Ethical considerations}

This work studies inference efficiency using existing models and public evaluation data and introduces no new human-subject data collection. Deployments should follow the applicable model and dataset licenses and isolate online prediction memories at appropriate user or service boundaries. Faster inference does not remove the need to assess outputs for the intended application.

\bibliography{recap}

\appendix

\section{Implementation and Baseline Settings}
\label{app:implementation}

\paragraph{Model execution.}
Qwen and LLaVA run in bfloat16 and float16, respectively, on the workloads
of Section~\ref{sec:setup}.
Timing covers prefill and decoding, excluding data loading and report aggregation.

\paragraph{\method\ defaults.}
We use Top-$8$ rows, width $8$, depth $10$, and $63$ transition candidates
excluding the root. Available rows receive priors $(0.70,0.20,0.10)$ by
decreasing specificity; entropy-confidence weighting multiplies each prior
by the row's confidence and normalizes the products
(Equation~\ref{eq:confidence-weights}). Depth priors are
$(0.68,0.63,0.50,0.43,0.40,0.38)$, then $0.38$. Prefix-sharing suffix
insertion adds at most four tokens beyond the transition cap, giving at
most $68$ packed nodes including the root; controls described as
suffix-inclusive instead count suffix nodes within the $63$-node cap.
Special image tokens and duplicate children are filtered.

\paragraph{Memory updates and packed verification.}
Prefill seeds $\U$ from each eligible prompt token's first occurrence;
$\C/\G$ start empty. All evaluated rows refresh memory, retaining each
key's first batch occurrence and replacing older entries.
Packed positions follow path depth; accepted-path key/value entries
continue decoding. Stream and policy states are isolated and reset after warmup;
in the main results, $\U$ is also cleared at each benchmark boundary, so no
predictions carry over between benchmarks.

\paragraph{Baselines.}
PLD uses $4$-token keys and at most $10$ draft tokens; Lookahead uses
decoding/branch lengths $64/12$. Token Recycling follows the official
implementation \citep{luo-etal-2025-turning}: a zero-initialized Top-$8$
adjacency matrix fills the static imbalanced ``2.2.2'' tree ($80$
candidates, depth $6$) layer by layer, and every verified node writes its
Top-$8$ target predictions back to the matrix. Prefill logits are not
written; the matrix persists across requests and is cleared only after
warmup and at benchmark boundaries. Tree Recycling expands a static tree
of width $2$ and depth $4$, giving $30$ candidates under a $31$-node cap,
from a request-local Top-$4$ matrix.
SAM uses total-token setting $64$ and draft lengths $2$--$40$.
SAM+ViSkip caps drafts at $40$ tokens and gates at $0.35$ on head-averaged
image attention at final-layer query zero.

\section{Timing and Analysis Protocols}
\label{app:protocol}

\paragraph{Fusion rule.}
The main results (Tables~\ref{tab:main} and~\ref{tab:mmspec}) and the
MMSpec latency analysis (Appendix~\ref{app:mmspec-latency}) use
entropy-confidence weighting. Component analyses in
Section~\ref{sec:analysis} and Appendices~\ref{app:additional},
\ref{app:scope}, \ref{app:overhead}, and~\ref{app:root-retrieval} use its
fixed-prior special case ($c_j=1$), so that each control varies one
component; Appendix~\ref{app:adaptive-fusion} compares the two rules directly.

\paragraph{Pairing and timed boundaries.}
For the recycling methods (\method, Tree Recycling, and Token Recycling),
SAM, and SAM+ViSkip, autoregressive references are interleaved with method runs; PLD and Lookahead pair generation units. Main methods each have a separate run and reference, while component controls share a run. Main recycling and MMSpec ablations synchronize at target forwards. SAM+ViSkip and four-dataset controls synchronize at generation boundaries; the four-dataset controls also share timing paths that include CPU candidate counters.

\paragraph{MMSpec controls.}
In MMSpec ablations, the ratio for input $j$ pools its multi-turn records:
\begin{equation}
 R_j^{a/b}=
 \frac{\sum_{i\in j}n_i^a/\sum_{i\in j}T_i^a}
 {\sum_{i\in j}n_i^b/\sum_{i\in j}T_i^b}.
\end{equation}
Gains are reported as $100(\operatorname{mean}_jR_j^{a/b}-1)$. Resampling and repeated request orders reuse the same inputs. Their variation therefore reflects this fixed evaluation set, not independent-dataset uncertainty.

\paragraph{Offline replay.}
Offline replay queries before updating, includes unmatched empty-memory states, and compares proposals with saved native target continuations. Trees respect a $63$-node cap and are rebuilt to the smallest actual count among policies. Matched length excludes the root and target correction token. Suffix additions are included in the cap for controls; the default configuration allows four extra nodes.

Offline means weight replay states within each dataset and use $2{,}000$ paired image-cluster resamples: $97$ ChartQA clusters and $100$ clusters elsewhere. Online resampling pairs inputs, using $100$ per dataset or the first $50$ for scope controls.

\section{Candidate-Level Results}
\label{app:additional}

Table~\ref{tab:granularity} gives the coverage and recall values
in Figure~\ref{fig:granularity}; the corresponding online gains are reported below.

\begin{table}[!t]
\centering
\small
\renewcommand{\arraystretch}{1.10}
\setlength{\tabcolsep}{3.5pt}
\begin{tabular*}{\columnwidth}{@{\extracolsep{\fill}}lrrrrrr@{}}
\toprule
\multirow{2}{*}{Dataset} & \multicolumn{3}{c}{Coverage (\%, $\uparrow$)} & \multicolumn{3}{c}{Recall@$8$ (\%, $\uparrow$)} \\
\cmidrule(lr){2-4}\cmidrule(lr){5-7}
 & $\U$ & $\C$ & $\G$ & $\U$ & $\C$ & $\G$ \\
\midrule
ChartQA & 44.5 & 23.2 & 14.8 & 64.5 & 81.7 & 86.6 \\
TextVQA & 26.3 & 7.4 & 3.4 & 52.5 & 73.5 & 78.3 \\
OCRBench & 54.9 & 30.4 & 18.6 & 63.6 & 84.2 & 90.6 \\
MMT-Bench & 45.6 & 19.7 & 11.2 & 56.9 & 77.9 & 84.4 \\
\bottomrule
\end{tabular*}
\caption{Coverage and conditional Recall@$8$ during empty-initialization replay. Queries are evaluated before their corresponding memory updates.}
\label{tab:granularity}
\end{table}

With $100$ paired inputs per dataset, shared initialization, and a
$63$-node cap, $\U+\C$ backoff improves throughput over $\U$ by
$2.29\%$, $1.11\%$, $3.64\%$, and $1.65\%$ on ChartQA, TextVQA,
OCRBench, and MMT-Bench, respectively.

\paragraph{Backoff versus fusion.}
Matched-node comparisons favor backoff over fusion on ChartQA and OCRBench, with similar continuation lengths elsewhere (Table~\ref{tab:path}).

\begin{table}[!t]
\centering
\small
\renewcommand{\arraystretch}{1.10}
\setlength{\tabcolsep}{3.5pt}
\begin{tabular*}{\columnwidth}{@{\extracolsep{\fill}}lrrrr@{}}
\toprule
\multirow{2}{*}{Policy} & \multicolumn{4}{c}{Matched continuation length ($\uparrow$)} \\
\cmidrule(lr){2-5}
 & ChartQA & TextVQA & OCR & MMT \\
\midrule
$\U$ & 0.516 & 0.224 & 0.632 & 0.444 \\
$\U+\C$ backoff & 0.709 & 0.253 & 1.135 & 0.600 \\
$\U+\C+\G$ backoff & 0.778 & 0.252 & 1.128 & 0.638 \\
Uniform fusion & 0.737 & 0.253 & 1.041 & 0.629 \\
Prior-weighted fusion & 0.751 & 0.253 & 1.066 & 0.639 \\
\bottomrule
\end{tabular*}
\caption{Matched continuation length in empty-initialization replay. Each state uses equal actual node counts across the compared policies. Prior-weighted fusion uses fixed priors ($c_j=1$).}
\label{tab:path}
\end{table}

\begin{table*}[!t]
\centering
\small
\renewcommand{\arraystretch}{1.10}
\setlength{\tabcolsep}{7pt}
\begin{tabular*}{\textwidth}{@{\extracolsep{\fill}}lrrrrr@{}}
\toprule
\multirow{2}{*}{Dataset} & \multicolumn{3}{c}{Offline matched continuation length} & \multicolumn{2}{c}{Online fusion + suffix / fusion} \\
\cmidrule(lr){2-4}\cmidrule(lr){5-6}
 & Fusion & + Suffix & Difference & Gain (\%, $\uparrow$) & $\tau$: before $\to$ after \\
\midrule
ChartQA & 1.123 & 1.313 & 0.190 & $+0.83$ & $2.80\to2.88$ \\
TextVQA & 0.355 & 0.413 & 0.059 & $+0.07$ & $2.18\to2.18$ \\
OCRBench & 1.417 & 1.538 & 0.120 & $-0.67$ & $2.66\to2.62$ \\
MMT-Bench & 0.750 & 0.866 & 0.116 & $+2.23$ & $2.40\to2.44$ \\
\bottomrule
\end{tabular*}
\caption{Suffix-retrieval controls. Offline matching omits root and correction tokens; online gains and acceptance length $\tau$ use Equations~\ref{eq:speed}--\ref{eq:acceptance}.}
\label{tab:suffix}
\end{table*}

\paragraph{Fusion in online controls.}
With empty initialization and fixed priors, prior-weighted/uniform throughput ratios are $1.0043$, $1.0015$, $1.0006$, and $1.0037$ on ChartQA, TextVQA, OCRBench, and MMT-Bench. These near-neutral ratios qualify the gain observed in the prompt-seeded MMSpec control (Table~\ref{tab:ablation}) and show that specificity priors have setting-dependent benefits.

\paragraph{Suffix complementarity.}
At equal actual candidate counts under a $63$-node cap, prompt-seeded suffix insertion adds first-token hits in $3.33\%$, $1.46\%$, $1.95\%$, and $2.31\%$ of states (Table~\ref{tab:suffix}). Online policies are interleaved, bridge extension is disabled, and all suffix-related operations are timed.

\paragraph{Image-conditioned source predictions.}
Holding queries fixed and using eight candidates per source, jointly available matching-image rows increase fused Recall@$8$ over swapped-image rows by $5.55$, $6.35$, $1.91$, and $0.89$ points on ChartQA, TextVQA, OCRBench, and MMT-Bench. Donor images are category-matched on ChartQA/OCRBench and pooled on the other datasets. The context-only/trigram-only changes of $-2.96/-2.50$ points on OCRBench show that longer-key sources do not benefit uniformly. No image-similarity gate is used.

\section{Prediction Reuse across Request Streams}
\label{app:scope}
\label{sec:scope}

\begin{figure}[!t]
\centering
\includegraphics[width=\columnwidth]{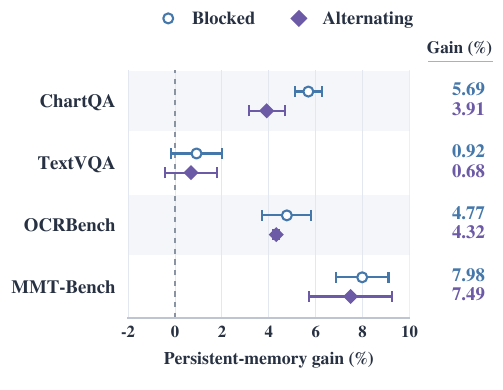}
\caption{Token-level reuse across task schedules. Persistent $\U$ is paired with per-request reset on identical inputs to measure throughput under blocked and alternating benchmark streams.}
\label{fig:scope}
\end{figure}

We test cross-task reuse by changing token-memory lifetime while holding contextual memory request-local. Persistent $\U$ is compared with per-request reset, and $\C$ and $\G$ reset in both cases. The same $200$ inputs are scheduled either in benchmark blocks or by alternating benchmarks.

Persistent memory improves throughput on ChartQA, OCRBench, and MMT-Bench under both schedules, with near-neutral effects on TextVQA (Figure~\ref{fig:scope}). MMT-Bench gains are $7.98\%$ for blocked streams and $7.49\%$ for alternating streams. A task-boundary reset retains reuse within benchmarks but clears $\U$ on each benchmark switch. It becomes equivalent to per-request reset under alternation, while persistent $\U$ retains predictions through intervening tasks.

\begin{table}[!t]
\centering
\small
\renewcommand{\arraystretch}{1.10}
\setlength{\tabcolsep}{6pt}
\begin{tabular*}{\columnwidth}{@{\extracolsep{\fill}}lrr@{}}
\toprule
\multirow{2}{*}{Dataset} & \multicolumn{2}{c}{Throughput gain (\%, $\uparrow$)} \\
\cmidrule(lr){2-3}
 & Blocked & Alternating \\
\midrule
ChartQA & $5.69\pm0.59$ & $3.91\pm0.77$ \\
TextVQA & $0.92\pm1.10$ & $0.68\pm1.13$ \\
OCRBench & $4.77\pm1.04$ & $4.32\pm0.13$ \\
MMT-Bench & $7.98\pm1.12$ & $7.49\pm1.77$ \\
\bottomrule
\end{tabular*}
\caption{Mean throughput gain $\pm$ sample SD for persistent $\U$ versus reset, measured across three orders of the same $200$ inputs.}
\label{tab:scope}
\end{table}

Within-dataset orders use seeds $17,29,43$. The three memory policies maintain isolated states, reset $\C/\G$ per request, and omit suffix insertion.

Table~\ref{tab:scope} averages paired input ratios separately for each seed. TextVQA's pooled ratios are $0.987$ under blocked scheduling and $0.990$ under alternation, indicating near-neutral effects sensitive to aggregation. Alternating task-reset and per-request-reset acceptance is identical across $600$ pairs. Memory snapshots include final prediction objects and exclude peak process RAM.

The persistent map keeps only the latest Top-$K$ row for each token, bounding $\U$ storage by $O(|\mathcal V|K)$ per stream. On cyclic replay of a fixed Qwen trajectory pool, its host footprint remains near $2.67$ MiB through $10{,}000$ request updates.

\section{Motivation Figure Diagnostics}
\label{app:intro-diagnostics}

Figure~\ref{fig:intro} uses Qwen2.5-VL-7B traces from $100$ images per single-turn benchmark, totaling $700$. We compute equal-benchmark averages and $10{,}000$ within-benchmark image-cluster bootstrap resamples. The counterfactuals hold text prefixes fixed while recomputing representations with mean-image and matched wrong-image controls.

\paragraph{Accepted length and root sensitivity (left).}
Visual dependence is the smaller counterfactual decrease in the original root token's log probability, and change rate records Top-$1$ changes under either control. Valid full-image anchors yield $15{,}677$ drafts across $695$ image clusters, grouped by accepted length into $0$--$4$ and $5+$. From rejection to $5+$ acceptance, dependence declines $0.620\to0.198$ and root-change rate declines $34.9\%\to12.4\%$.

\paragraph{Candidate-source alignment (right).}
For token-only candidates $C_U$ and contextual candidates $C_{GC}$, define the log candidate mass and visual support as
\begin{align}
 L_s(J)&=\log\sum_{z\in C_s}P_\theta(z\mid J,p,y_{<t}),\\
 E_s&=\min_{J\in\{I_{\mathrm{mean}},I_{\mathrm{wrong}}\}}
             \bigl[L_s(I)-L_s(J)\bigr].
\end{align}
We retain $1{,}125$ eligible states with valid full-image anchors, both sources, row overlap of at most $25\%$, and matched budgets of $2$--$8$ candidates. Within-benchmark quintiles of $E_{GC}-E_U$ run from $U$-aligned Q1 to context-aligned Q5. Bars show $\exp(E_s)/(\exp(E_U)+\exp(E_{GC}))$. The corresponding contextual acceptance gains are $-1.292$, $-0.375$, $0.709$, $1.239$, and $2.084$ tokens. Acceptance is not used to define the bins.

\begin{table}[!t]
\centering
\small
\setlength{\tabcolsep}{2.5pt}
\begin{tabular*}{\columnwidth}{@{\extracolsep{\fill}}lrrrr@{}}
\toprule
 & \multicolumn{2}{c}{Qwen} & \multicolumn{2}{c}{LLaVA} \\
Stage & Wall & CPU & Wall & CPU \\
\midrule
Suffix lookup & $0.010$ & $0.016$ & $0.010$ & $0.016$ \\
Lookup, fusion, search & $0.506$ & $0.510$ & $0.592$ & $0.598$ \\
Packing & $0.391$ & $0.409$ & $0.519$ & $0.538$ \\
Memory refresh & $0.461$ & $0.468$ & $0.414$ & $0.421$ \\
Suffix index update & $0.027$ & $0.034$ & $0.028$ & $0.034$ \\
Token commitment & $0.256$ & $0.273$ & $0.286$ & $0.304$ \\
Other decoding & $0.127$ & $0.148$ & $0.125$ & $0.145$ \\
Target verification & $19.912$ & $19.928$ & $33.241$ & $33.265$ \\
\bottomrule
\end{tabular*}
\caption{Decoding-phase times in milliseconds, pooled over iterations at a $63$-node cap with $40$ inputs per model and two orders. Wall times are exclusive; CPU time includes synchronization polling.}
\label{tab:overhead-stages}
\end{table}

\begin{table*}[!t]
\centering
\small
\begin{tabular*}{\textwidth}{@{\extracolsep{\fill}}llrrrrr@{}}
\toprule
Model & Cap & Mean nodes & Prep. ms & $p_{95}$ ms & Host MiB & GPU GiB \\
\midrule
Qwen & 31 & 26.5 & 0.591 & 0.722 & 5.79 & 19.32 \\
Qwen & 63 & 55.2 & 0.907 & 1.227 & 7.88 & 19.32 \\
Qwen & 127 & 113.4 & 2.202 & 2.617 & 8.74 & 19.32 \\
\midrule
LLaVA & 31 & 29.1 & 0.721 & 0.882 & 7.46 & 19.98 \\
LLaVA & 63 & 59.2 & 1.121 & 1.385 & 8.70 & 20.01 \\
LLaVA & 127 & 121.0 & 2.485 & 2.729 & 12.68 & 20.16 \\
\bottomrule
\end{tabular*}
\caption{Budget sensitivity with shared inputs and orders. Candidate counts include suffixes and exclude the root. Preparation covers lookup through packing; $p_{95}$ is the request-average 95th percentile. Host memory is the largest proposal-state snapshot. GPU memory is peak model-process allocation, including weights and generation workspaces.}
\label{tab:overhead-budgets}
\end{table*}

\section{Properties of Prediction Reuse}
\label{app:theory}

\paragraph{Context refinement.}
Let $Z$ be the greedy target token under fixed tie-breaking and $H_j$
the last $j$ history tokens, start-padded as needed. Optimal recall is
\begin{equation}
 R_K^*(H_j)=\mathbb E\!\left[
 \max_{\substack{\mathcal S\subseteq\mathcal V\\|\mathcal S|\le K}}
 \Pr(Z\in\mathcal S\mid H_j)\right].
 \label{eq:oracle-recall}
\end{equation}
For $H_a=f(H_b)$, any coarse selector can be reproduced as
$\mathcal S_b(H_b)=\mathcal S_a(f(H_b))$, proving $R_K^*(H_a)\le R_K^*(H_b)$.
Dropping the earliest token gives Equation~\ref{eq:context-refinement}.
Finite memories still require shorter-key coverage.

\paragraph{Fusion and sparse selection.}
At a fixed query, let $m=\sum_jw_jP_j$ combine normalized rows
$P_j=\widehat P_{M_j}$ with fixed positive weights summing to one.
The KL decomposition
\begin{align}
 &\sum_jw_jD_{\mathrm{KL}}(P_j\|\rho)\nonumber\\
 &\quad=\sum_jw_jD_{\mathrm{KL}}(P_j\|m)
       +D_{\mathrm{KL}}(m\|\rho).
 \label{eq:kl-decomposition}
\end{align}
makes $m$ the unique minimizer in Equation~\ref{eq:fusion-centroid},
without row independence \citep{Abbas_2009}.

Let $\mathcal S_K$ retain the largest $K$ positive masses of $m$ (all if fewer),
with deterministic ties, and $q=m(\cdot\mid\mathcal S_K)$. Then
\begin{equation}
 q\in\underset{\substack{\rho\in\Delta(\mathcal V)\\
                |\operatorname{supp}(\rho)|\le K}}{\arg\min}
 D_{\mathrm{KL}}(\rho\|m).
 \label{eq:sparse-projection}
\end{equation}
For nonempty $\mathcal S\subseteq\operatorname{supp}(m)$ and $\rho$ supported on $\mathcal S$,
\begin{align}
 D_{\mathrm{KL}}(\rho\|m)
 &=D_{\mathrm{KL}}\!\left(\rho\|m(\cdot\mid\mathcal S)\right)\nonumber\\
 &\quad-\log m(\mathcal S).
 \label{eq:support-decomposition}
\end{align}
Thus $\rho=m(\cdot\mid\mathcal S)$ attains $-\log m(\mathcal S)$, and Top-$K$
maximizes retained mass $m(\mathcal S)$, allowing tied optimal supports.
These properties characterize proposals; target verification determines commitment.

\section{Overhead and Memory Scaling}
\label{app:overhead}

\begin{table}[!t]
\centering
\small
\begin{tabular*}{\columnwidth}{@{\extracolsep{\fill}}llrr@{}}
\toprule
Model & $|\U|$ & Host MiB & Lookup ns \\
\midrule
Qwen & $1{,}000$ & $0.67$ & $109$ \\
 & $10{,}000$ & $7.03$ & $141$ \\
 & $152{,}064$ & $109.14$ & $386$ \\
\midrule
LLaVA & $1{,}000$ & $0.67$ & $109$ \\
 & $10{,}000$ & $7.03$ & $140$ \\
 & $32{,}064$ & $23.36$ & $183$ \\
\bottomrule
\end{tabular*}
\caption{Synthetic occupancy stress test: host-object storage and median lookup latency, measured over seven repeats of $100{,}000$ seeded ID/score hits with a checksum.}
\label{tab:memory-capacity}
\end{table}

\begin{table}[!t]
\centering
\small
\renewcommand{\arraystretch}{1.10}
\setlength{\tabcolsep}{4pt}
\begin{tabular*}{\columnwidth}{@{\extracolsep{\fill}}lrr@{}}
\toprule
Variant & $S$ ($\times$, $\uparrow$) & $\tau$ ($\uparrow$) \\
\midrule
\method\ (lexical) & \textbf{1.63} & 2.67 \\
\method\ + semantic fallback & 1.61 & \textbf{2.78} \\
\method\ + recency fallback & 1.61 & 2.75 \\
\bottomrule
\end{tabular*}
\caption{Root fallback on MMSpec with Qwen2.5-VL-7B. All variants use fixed-prior fusion ($c_j=1$). Longer accepted continuations accompany lower speedup. Best column values are bold.}
\label{tab:root-retrieval}
\end{table}

\begin{table*}[!t]
\centering
\small
\setlength{\tabcolsep}{2pt}
\begin{tabular*}{\textwidth}{@{\extracolsep{\fill}}l*{12}{r}@{}}
\toprule
\multirow{2}{*}{Variant} & \multicolumn{6}{c}{Qwen} & \multicolumn{6}{c}{LLaVA} \\
\cmidrule(lr){2-7}\cmidrule(lr){8-13}
 & ChartQA & TextVQA & OCR & MMT & Mean & $\tau$ & ChartQA & TextVQA & OCR & MMT & Mean & $\tau$ \\
\midrule
\rowcolor{recaprow}
\method & $+0.65$ & $+1.24$ & $+1.43$ & $-0.41$ & $+0.73$ & $2.93$ & $+2.30$ & $+0.34$ & $+1.71$ & $+0.01$ & $+1.09$ & $2.90$ \\
w/o confidence & $+0.00$ & $+0.00$ & $+0.00$ & $+0.00$ & $+0.00$ & $2.91$ & $+0.00$ & $+0.00$ & $+0.00$ & $+0.00$ & $+0.00$ & $2.88$ \\
w/o prior & $+0.37$ & $+0.56$ & $+1.38$ & $-0.93$ & $+0.35$ & $2.91$ & $+0.26$ & $-0.67$ & $+0.31$ & $-0.92$ & $-0.25$ & $2.85$ \\
Uniform & $+0.01$ & $+0.37$ & $+2.34$ & $+0.22$ & $+0.73$ & $2.90$ & $-0.47$ & $+0.18$ & $+2.29$ & $-0.06$ & $+0.48$ & $2.86$ \\
Confidence gate & $-1.10$ & $+1.52$ & $-1.33$ & $-1.72$ & $-0.66$ & $2.83$ & $-2.27$ & $-1.56$ & $-1.25$ & $-2.53$ & $-1.90$ & $2.76$ \\
\bottomrule
\end{tabular*}
\caption{Online fusion ablations, reporting mean input-paired throughput changes (\%) against w/o confidence and acceptance length $\tau$. Each model processes $200$ inputs, $50$ per dataset, under two orders. Policies share memory scope, candidate limits, and suffix insertion; timing includes weighting. Full \method\ is shaded.}
\label{tab:adaptive-fusion-online}
\end{table*}

\begin{table}[!t]
\centering
\small
\setlength{\tabcolsep}{2pt}
\begin{tabular*}{\columnwidth}{@{\extracolsep{\fill}}lrrrr@{}}
\toprule
Variant & ChartQA & TextVQA & OCR & MMT \\
\midrule
\rowcolor{recaprow}
\method & $+0.00$ & $+0.00$ & $+0.01$ & $-0.01$ \\
w/o confidence & $+0.00$ & $+0.00$ & $+0.00$ & $+0.00$ \\
w/o prior & $-0.08$ & $+0.00$ & $-0.31$ & $+0.00$ \\
Uniform & $-0.08$ & $+0.00$ & $-0.30$ & $+0.02$ \\
Confidence gate & $-1.49$ & $-0.14$ & $-1.99$ & $-0.95$ \\
\midrule
Specificity backoff & $-0.11$ & $+0.00$ & $+1.29$ & $-0.24$ \\
w/o conf., soft prior & $-0.04$ & $+0.00$ & $-0.16$ & $+0.00$ \\
w/o conf., sharp prior & $-0.02$ & $+0.03$ & $+0.15$ & $-0.05$ \\
\bottomrule
\end{tabular*}
\caption{Changes in all-query Recall@$8$ relative to w/o confidence (percentage points), evaluated on $23{,}811$ Qwen states from $400$ inputs. Uncovered states count as misses. Shading identifies \method.}
\label{tab:adaptive-fusion-offline}
\end{table}

\subsection{Latency and Workspace}

Profiling uses the same A100-SXM4-80GB GPU and AMD EPYC 9654 CPU for both models, with PyTorch $2.7.1$, CUDA $12.6$, and four CPU threads. Ten inputs per diagnostic dataset follow Appendix~\ref{app:implementation}. Orders $17/29$ interleave caps $31/63/127$, resetting isolated states after warmup and between repeats. Profiling uses the fixed-prior special case ($c_j=1$); entropy-confidence weighting additionally evaluates one row entropy per retrieved row at multi-source queries. In an isolated host microbenchmark of the reference Python fusion over $512$ multi-source queries, this raises fusion time from $7.3$ to $14.5\,\mu$s per query; the C++ builder used for the main results was not profiled separately.

We pool exclusive CUDA-synchronized phases over decoding iterations, excluding prefill and image encoding (Table~\ref{tab:overhead-stages}). Packing includes topology, masks, positions, and verification inputs. Memory traversal and serialization are untimed. An uninstrumented $63$-node stream measures instrumentation overhead, and four $64$-token traces per model separately attribute CUDA kernel and memory-operation time.

\paragraph{Stage costs.}
Preparation takes $0.907/1.121$ ms on Qwen/LLaVA, or $4.18\%/3.18\%$ of decoding wall time. Adding memory refresh and suffix-index updates gives $1.395/1.563$ ms ($6.43\%/4.44\%$), and all stages outside target verification total $1.778/1.974$ ms ($8.20\%/5.61\%$). Instrumentation adds $1.01\%/1.36\%$ median latency. Retrieval, fusion, and search launch no CUDA work. Packing consumes $4.46/3.29\,\mu$s of device time per iteration, while verification consumes $16.60/28.04$ ms, representing $98.88\%/99.61\%$ of decoding-device time.

\paragraph{Budget sensitivity and workspace.}
For $B$ candidates, the ancestor submatrix contains $(B+1)^2$ entries. Table~\ref{tab:overhead-budgets} deduplicates host objects shared across $\U/\C/\G$, the suffix index, and topology state.

\subsection{Persistent-Memory Growth}

One latest Top-$K$ row per token bounds persistent memory by $|\U|\leq|\mathcal V|$. The resulting space cost is $O(|\mathcal V|K)$ per stream and $O(n_s|\mathcal V|K)$ for $n_s$ streams, independent of request count. Request-local $\C/\G$ scale with the contexts evaluated within each request.

\begin{figure}[!t]
\centering
\includegraphics[width=\columnwidth]{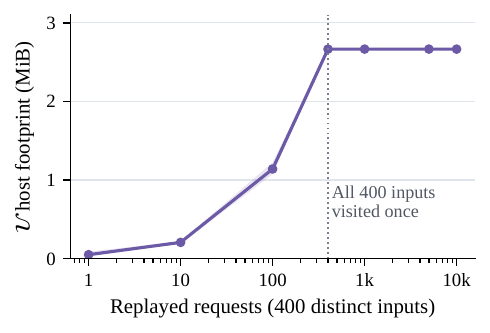}
\caption{Persistent $\U$ over cyclic replay of the same $400$ trajectories. Curves report the mean and min--max range across three replay orders.}
\label{fig:memory-scaling}
\end{figure}

\paragraph{Replay of observed prediction rows.}
Cyclic replay reshuffles $400$ native Qwen trajectories, $100$ per diagnostic dataset, under seeds $17/29/43$ at each pass. Top-$8$ memory reaches $3{,}821$ keys and $2.67$ MiB after $400$ requests and stays stable through $10{,}000$ updates (Figure~\ref{fig:memory-scaling}). This result characterizes replacement on a fixed pool, not saturation under unseen inputs.

\paragraph{Vocabulary-capacity stress test.}
The capacity stress test fills synthetic Top-$8$ rows at three occupancies in fresh processes (Table~\ref{tab:memory-capacity}). At full vocabulary occupancy, Qwen/LLaVA host storage is $109.14/23.36$ MiB and RSS increases by $126.48/27.55$ MiB. These capacity figures do not represent natural-stream footprints. Persistent rows contain no CUDA tensors.

Across all six occupancies, $63$-node tree construction takes $0.242$--$0.249$ ms, measured as the median of seven repeats of $128$ queries with packing excluded.

\begin{figure*}[!t]
\centering
\includegraphics[width=\textwidth]{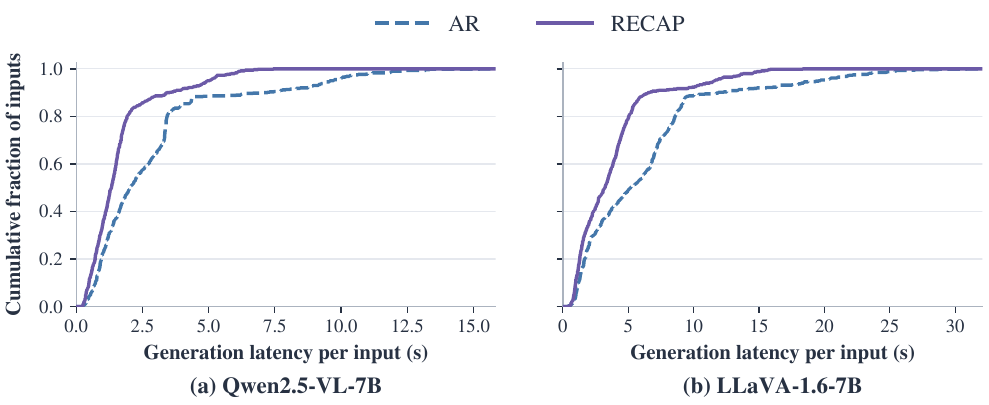}
\caption{Per-input generation-latency CDFs on MMSpec for paired \method\ and AR runs. A leftward curve indicates lower latency. Multi-turn inputs contribute summed generation time across turns.}
\label{fig:mmspec-latency}
\end{figure*}

\begin{table}[!t]
\centering
\small
\renewcommand{\arraystretch}{1.12}
\setlength{\tabcolsep}{2pt}
\begin{tabular*}{\columnwidth}{@{\extracolsep{\fill}}llrrrr@{}}
\toprule
Model & Method & P50 & P90 & P95 & P99 \\
\midrule
Qwen & AR & 2.70 & 8.44 & 11.63 & 15.07 \\
\rowcolor{recaprow}
Qwen & \method & \textbf{1.52} & \textbf{4.46} & \textbf{6.08} & \textbf{7.56} \\
\midrule
LLaVA & AR & 4.02 & 9.42 & 14.27 & 18.43 \\
\rowcolor{recaprow}
LLaVA & \method & \textbf{2.38} & \textbf{5.45} & \textbf{8.25} & \textbf{10.67} \\
\bottomrule
\end{tabular*}
\caption{MMSpec per-input generation-latency percentiles in seconds, with lower values indicating faster generation. P$k$ denotes the $k$th percentile; \method\ rows are shaded.}
\label{tab:mmspec-latency}
\end{table}

\section{Alternative Root-Level Retrieval}
\label{app:root-retrieval}

Qwen fallbacks use a fixed six-category MMSpec subset and activate only
when all root lexical lookups fail; descendants retain lexical retrieval.
Semantic fallback selects up to three rows by cosine similarity of
verified and predecessor states, with no extra target forwards.
A request-local FIFO index holds at most $2{,}048$ surviving $\U/\C/\G$
rows; recency fallback selects the latest.
Both use fixed priors $(0.70,0.20,0.10)$ without confidence ($c_j=1$)
and Top-$8$ candidates.

Interleaved policies share target-generated histories, depth $10$,
a $63$-candidate cap including suffixes but excluding the root,
and isolated persistent $\U$ and request-local $\C/\G$.
Timing includes image encoding, prefill, index maintenance, and decoding;
metrics average turns.

Both increase $\tau$ but reduce speedup by $1.46\%$ (semantic)
and $1.34\%$ (recency; Table~\ref{tab:root-retrieval}).
Semantic lookup activates in $4.87\%$ of iterations despite continuous
index maintenance. Neither fallback is enabled by default.

\begin{figure*}[!t]
\centering
\includegraphics[width=0.48\textwidth]{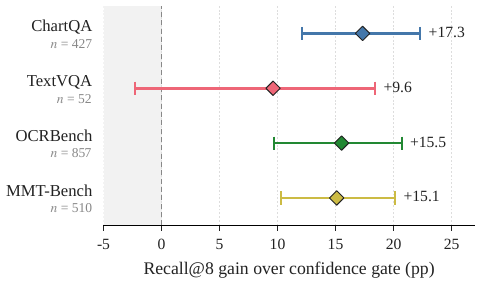}
\caption{Fusion gains over confidence gating on Qwen replay states with
source disagreement. Points show Recall@$8$ differences; bars give $95\%$
confidence intervals from $2{,}000$ paired image-cluster bootstrap resamples.
The dashed line marks zero gain, and $n$ is the number of disagreement states.}
\label{fig:fusion-disagreement}
\end{figure*}

\begin{figure*}[!t]
\centering
\includegraphics[width=0.92\textwidth]{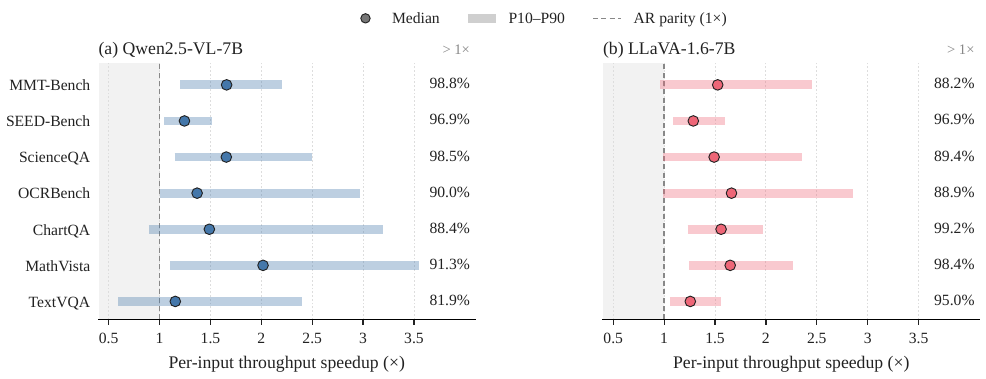}
\caption{Per-input throughput speedup on the seven single-turn benchmarks
($1{,}000$ inputs per model and benchmark). Points mark medians and segments
span P10--P90 of the input-level ratios. Dashed lines mark parity with AR;
the percentages report inputs with strictly higher throughput than paired AR.}
\label{fig:single-speedup-distribution}
\end{figure*}

\section{Fusion Ablations}
\label{app:adaptive-fusion}

In Equation~\ref{eq:confidence-weights}, \emph{w/o confidence}
sets $c_j(h)=1$ (reference), \emph{w/o prior} sets $\alpha_j=1$,
and uniform fusion removes both factors. Hard gating selects the
most confident row, breaking ties by specificity.
All rules share retrieved rows, Top-$8$ truncation, and renormalization.

\paragraph{Offline matching.}
Replay uses earlier rows, prompt-seeded $\U$, and empty $\C/\G$.
\method\ changes all-query Recall@$8$ by less than $0.02$ points;
gating loses $0.14$--$1.99$ points
(Table~\ref{tab:adaptive-fusion-offline}).
Without confidence, soft $(0.55,0.30,0.15)$ and sharp $(0.85,0.10,0.05)$
priors change recall by at most $0.17$ points.

\paragraph{Online comparison.}
The $4{,}000$ generations in Table~\ref{tab:adaptive-fusion-online}
compare five rules with isolated memories, suffix insertion,
and Appendix~\ref{app:implementation}'s settings.
Policies rotate per request; timing includes image encoding,
prefill, generation, and adaptation. Repeats reuse inputs to assess order variation.
\method\ gains $0.73\%/1.09\%$ mean throughput on Qwen/LLaVA,
with median paired changes of $-0.12\%/-0.07\%$.
W/o prior (confidence only) and uniform fusion have mixed effects; gating lowers
throughput on both models. Gains are modest and task-dependent.
Confidence measures concentration; learned gates and visual-dependence
signals remain untested.

\paragraph{Disagreement among sources.}
We further examine replay states where at least two retrieved rows have
different Top-$1$ predictions. On these states, \method\ exceeds confidence
gating by $9.62$--$17.33$ Recall@$8$ points
(Figure~\ref{fig:fusion-disagreement}; Table~\ref{tab:fusion-disagreement}).
Paired image-cluster intervals exclude
zero on ChartQA, OCRBench, and MMT-Bench; TextVQA has only $52$ such states
and a wider interval crossing zero. These results support retaining
complementary candidates when sources disagree. They isolate the benefit
over hard source selection; the effect of confidence weighting relative
to fixed priors is assessed separately above.

\begin{table*}[!t]
\centering
\small
\renewcommand{\arraystretch}{1.10}
\begin{tabular*}{\textwidth}{@{\extracolsep{\fill}}lrrrrr@{}}
\toprule
Dataset & States & \method & Confidence gate & Difference & 95\% CI \\
\midrule
ChartQA & 427 & 84.31 & 66.98 & $+17.33$ & $[12.08, 22.31]$ \\
TextVQA & 52 & 75.00 & 65.38 & $+9.62$ & $[-2.33, 18.42]$ \\
OCRBench & 857 & 62.43 & 46.91 & $+15.52$ & $[9.69, 20.70]$ \\
MMT-Bench & 510 & 77.65 & 62.55 & $+15.10$ & $[10.27, 20.14]$ \\
\bottomrule
\end{tabular*}
\caption{Recall@$8$ (\%) on Qwen replay states with source disagreement.
Differences and confidence intervals are in percentage points. Intervals use
$2{,}000$ paired image-cluster bootstrap resamples of the fixed evaluation set.}
\label{tab:fusion-disagreement}
\end{table*}

\paragraph{Observed fusion weights.}
Table~\ref{tab:fusion-weight-activity} summarizes normalized weights from
the online \method\ arm, grouping calls by the number of available rows.
Each candidate-tree fusion call contributes equally, pooling both request
orders; indices follow decreasing specificity among available sources.
With three sources, pooled weights are $(0.693,0.199,0.108)$ for Qwen and
$(0.688,0.200,0.112)$ for LLaVA, close to the prior $(0.700,0.200,0.100)$.
With two sources, the normalized prior is $(0.778,0.222)$, compared with
observed means of $(0.764,0.236)$ and $(0.763,0.237)$.
Thus confidence adjusts the allocation while preserving the average
preference for specific sources. These averages describe aggregate
weighting behavior; they do not imply small adjustments at every query.

\begin{table*}[!t]
\centering
\small
\renewcommand{\arraystretch}{1.10}
\setlength{\tabcolsep}{3pt}
\begin{tabular*}{\textwidth}{@{\extracolsep{\fill}}llrlrl@{}}
\toprule
\multirow{2}{*}{Model} & \multirow{2}{*}{Dataset} & \multicolumn{2}{c}{Two available sources} & \multicolumn{2}{c}{Three available sources} \\
\cmidrule(lr){3-4}\cmidrule(lr){5-6}
 & & Calls & Mean $(w_1,w_2)$ & Calls & Mean $(w_1,w_2,w_3)$ \\
\midrule
Qwen & ChartQA & 15,101 & $(0.764, 0.236)$ & 12,072 & $(0.697, 0.197, 0.105)$ \\
 & TextVQA & 12,689 & $(0.772, 0.228)$ & 8,353 & $(0.700, 0.197, 0.103)$ \\
 & OCRBench & 20,883 & $(0.758, 0.242)$ & 19,394 & $(0.693, 0.199, 0.108)$ \\
 & MMT-Bench & 44,809 & $(0.765, 0.235)$ & 32,106 & $(0.689, 0.200, 0.110)$ \\
 & Pooled & 93,482 & $(0.764, 0.236)$ & 71,925 & $(0.693, 0.199, 0.108)$ \\
\midrule
LLaVA & ChartQA & 33,922 & $(0.764, 0.236)$ & 20,304 & $(0.694, 0.198, 0.108)$ \\
 & TextVQA & 17,023 & $(0.770, 0.230)$ & 6,563 & $(0.700, 0.199, 0.102)$ \\
 & OCRBench & 29,778 & $(0.759, 0.241)$ & 32,512 & $(0.687, 0.200, 0.113)$ \\
 & MMT-Bench & 44,589 & $(0.763, 0.237)$ & 36,050 & $(0.685, 0.201, 0.115)$ \\
 & Pooled & 125,312 & $(0.763, 0.237)$ & 95,429 & $(0.688, 0.200, 0.112)$ \\
\bottomrule
\end{tabular*}
\caption{Mean normalized fusion weights in the online ablation, averaged
over calls at all candidate-tree positions. Each model uses $200$ inputs
under two orders. Source indices denote specificity ranks among available
rows. Single-source calls have weight one and are omitted.}
\label{tab:fusion-weight-activity}
\end{table*}

\begin{table*}[!t]
\centering
\small
\renewcommand{\arraystretch}{1.10}
\setlength{\tabcolsep}{4pt}
\begin{tabular*}{\textwidth}{@{\extracolsep{\fill}}lrrrrrrrr@{}}
\toprule
\multirow{2}{*}{Dataset} & \multicolumn{4}{c}{Qwen} & \multicolumn{4}{c}{LLaVA} \\
\cmidrule(lr){2-5}\cmidrule(lr){6-9}
 & P10 & P50 & P90 & $r_i>1$ (\%) & P10 & P50 & P90 & $r_i>1$ (\%) \\
\midrule
MMT-Bench & 1.20 & 1.66 & 2.20 & 98.8 & 0.96 & 1.53 & 2.46 & 88.2 \\
SEED-Bench & 1.05 & 1.24 & 1.52 & 96.9 & 1.09 & 1.29 & 1.60 & 96.9 \\
ScienceQA & 1.15 & 1.66 & 2.50 & 98.5 & 0.99 & 1.49 & 2.36 & 89.4 \\
OCRBench & 1.00 & 1.37 & 2.97 & 90.0 & 0.99 & 1.66 & 2.86 & 88.9 \\
ChartQA & 0.89 & 1.49 & 3.19 & 88.4 & 1.23 & 1.56 & 1.97 & 99.2 \\
MathVista & 1.10 & 2.02 & 3.55 & 91.3 & 1.25 & 1.65 & 2.26 & 98.4 \\
TextVQA & 0.59 & 1.15 & 2.40 & 81.9 & 1.06 & 1.26 & 1.56 & 95.0 \\
\bottomrule
\end{tabular*}
\caption{Distributions of paired throughput ratios for \method\ on the
seven single-turn benchmarks ($1{,}000$ inputs per dataset and model).
P10/P50/P90 report speedup factors; the final column for each model gives
the percentage of inputs with strictly higher throughput than paired AR.}
\label{tab:single-speedup-distribution}
\end{table*}

\section{Per-Input Runtime Distributions}
\label{app:mmspec-latency}

\paragraph{MMSpec latency.}
Following MMSpec \citep{shen2026mmspecbenchmarkingspeculativedecoding},
we compare paired AR/\method\ generation times from Table~\ref{tab:mmspec}.
Input $i$ contributes $T_i^{(m)}=\sum_{r=1}^{R_i}t_{ir}^{(m)}$ over
$R_i$ turns. Each $t_{ir}^{(m)}$ includes prefill and decoding,
excluding prompt preparation and output text decoding.
All $600$ inputs enter
$\widehat F_m(t)=N^{-1}\sum_{i=1}^{N}\mathbf{1}[T_i^{(m)}\leq t]$
without filtering.

\method\ lowers median and upper-tail latency
(Figure~\ref{fig:mmspec-latency}; Table~\ref{tab:mmspec-latency}).
Qwen/LLaVA P95 falls by $47.69\%/42.17\%$,
and $87.33\%/97.83\%$ of inputs are faster than paired AR.

\paragraph{Single-turn speedup.}
For each single-turn benchmark in Table~\ref{tab:main}, we summarize all
$1{,}000$ paired throughput ratios from Equation~\ref{eq:speed}
without filtering (Figure~\ref{fig:single-speedup-distribution};
Table~\ref{tab:single-speedup-distribution}).
These are empirical input-level percentiles, not uncertainty intervals.
Median speedup exceeds one on every benchmark for both models, and
$81.9$--$98.8\%$ of Qwen inputs and $88.2$--$99.2\%$ of LLaVA inputs
have higher throughput than paired AR. The distributions nevertheless vary
across tasks: Qwen reaches a median of $1.66\times$ on MMT-Bench, with
$98.8\%$ of inputs benefiting, versus $1.15\times$ and $81.9\%$ on TextVQA.
The latter also has a P10 of $0.59\times$, showing that positive average
gains coexist with substantial slowdowns on some inputs.

\end{document}